\documentclass[preprint,12pt]{elsarticle}

\usepackage{amssymb}
\usepackage{amsmath}
\usepackage{booktabs}
\usepackage{multirow}
\usepackage{multicol}
\usepackage{amsmath}
\usepackage{bm}
\usepackage{etoolbox}
\usepackage{amssymb}
\usepackage{xcolor}     
\usepackage{colortbl}   
\usepackage{soul}

\definecolor{lightblue}{RGB}{220, 230, 245}

\journal{Nuclear Physics B}

\begin{document}

\begin{frontmatter}



\title{Vision-Language Semantic Aggregation Leveraging Foundation Model for Generalizable Medical Image Segmentation}


\author[1]{Wenjun Yu\fnref{fn1}}

\author[1]{Yinchen Zhou\fnref{fn1}}

\author[1]{Jia-Xuan Jiang\corref{cor1}}
\ead{jiangjx2023@lzu.edu.cn}
\author[1]{Shubin Zeng}
\ead{zenshubing@lzu.edu.cn}
\author[1]{Yuee Li}
\ead{liyuee@lzu.edu.cn}
\author[1]{Zhong Wang\corref{cor1}}
\ead{wangzhong@lzu.edu.cn}


\fntext[fn1]{Equal contribution to this research.}

\cortext[cor1]{Corresponding author}

\affiliation[1]{organization={School of Information Science \& Engineering, Lanzhou University},
    addressline={}, 
    city={Lanzhou},
    postcode={730000}, 
    country={China}}

\begin{abstract}
Multimodal models have achieved remarkable success in natural image segmentation, yet they often underperform when applied to the medical domain. Through extensive study, we attribute this performance gap to the challenges of multimodal fusion, primarily the significant semantic gap between abstract textual prompts and fine-grained medical visual features, as well as the resulting feature dispersion. To address these issues, we revisit the problem from the perspective of semantic aggregation. Specifically, we propose an Expectation-Maximization (EM) Aggregation mechanism and a Text-Guided Pixel Decoder. The former mitigates feature dispersion by dynamically clustering features into compact semantic centers to enhance cross-modal correspondence. The latter is designed to bridge the semantic gap by leveraging domain-invariant textual knowledge to effectively guide deep visual representations. The synergy between these two mechanisms significantly improves the model's generalization ability. Extensive experiments on public cardiac and fundus datasets demonstrate that our method consistently outperforms existing SOTA approaches across multiple domain generalization benchmarks.
\end{abstract}

\begin{highlights}
\item Proposes a vision-language aggregation framework for generalizable medical image segmentation 
\item Utilizes Expectation-Maximization (EM) to aggregate deep visual features, reducing dispersion
\item Designs a Text-Guided Pixel Decoder to enhance semantic understanding and generalization
\item Achieves state-of-the-art performance, demonstrating strong generalization capability
\end{highlights}

\end{frontmatter}



\section{Introduction}
Developing models that can robustly generalize to data collected across different medical institutions and imaging devices has long been a central goal and major challenge in medical image segmentation. To address this issue, researchers have proposed various methods. Some have utilized techniques like UDA\cite{wilson2020survey,li2024comprehensive} and MSDG\cite{khoee2024domain,shu2021open}, achieving decent results in specific scenarios by leveraging unlabeled target domain data or data from multiple diverse source domains for training. However, UDA\cite{wilson2020survey,li2024comprehensive} methods typically require access to target domain data (even if unlabeled), which is often impractical in real clinical deployment. While MSDG\cite{khoee2024domain,shu2021open} methods don't rely on target domain data\cite{Liu_Yin_Qu_Wang_2022}, their effectiveness is limited by the diversity and coverage of available source domain data, and they incur significantly higher training overhead. In contrast, Single-Source Domain Generalization (SDG)\cite{Zhou_Liu_Qiao_Xiang_Loy_2022,Xu_Xie_Reynolds_Ragoza_Gong_Batmanghelich} methods rely solely on data from a single source domain for training, aiming to learn more universal representations, offering lower deployment requirements and greater practical value. In recent years, several classes of SDG methods ,such as PCSDG\cite{jiang2025structure} and MRFFD\cite{wang2025multi}, including meta-learning, data augmentation, feature disentanglement, and regularization have made significant progress in enhancing model generalization capabilities. With the rise of Vision-Language Models (VLMs) and Large Language Models (LLMs)\cite{vaswani2017attention}, researchers have begun exploring their powerful cross-modal understanding and generalization abilities. Existing studies have attempted to integrate powerful pre-trained vision encoders (e.g., CLIP\cite{Radford_Kim_Hallacy_Ramesh_Goh_Agarwal_Sastry_Amanda_Mishkin_Clark_etal._2021}'s ViT) or leverage the rich semantic priors and contextual information provided by LLMs\cite{vaswani2017attention}A to enhance model generalization. These approaches have shown initial promise on several medical segmentation benchmarks. However,recent methods like TQDM\cite{pak2024textual} show limited performance on the domain generalization problem in medical image segmentation, for visual features (especially the low-level texture and structural information often relied upon in medical images) tend to be relatively shallow, while the semantic information provided by language models is deep-level and highly abstract. So how to make the effective alignment and fusion of these two types of information remains a core challenge\cite{yang2022unified}.

\begin{figure}[htbp]
  \centering
  \includegraphics[width=\columnwidth]{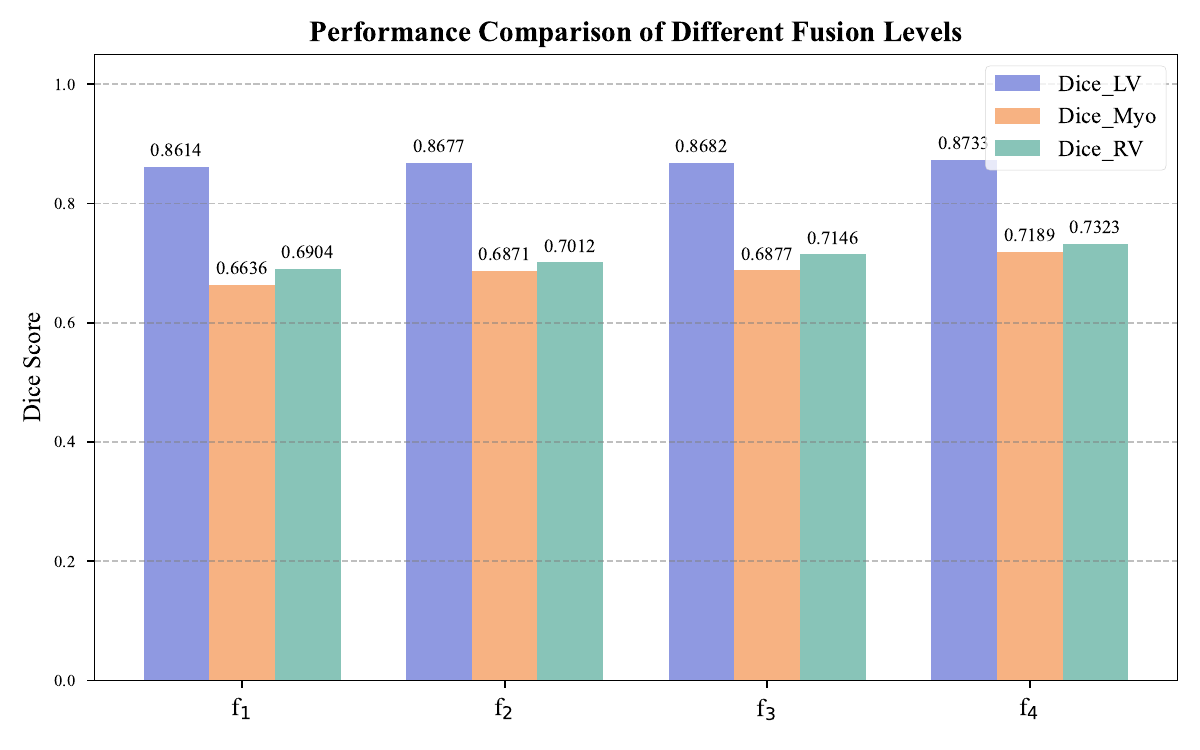}
  \caption{The bar chart compares the Dice Score on our samples for each cardiac structure when text-guided fusion is applied at different levels of the visual feature pyramid.}
  \label{fig:fusion_level_ablation}
\end{figure}

As shown in Fig.~\ref{fig:fusion_level_ablation}, extensive experiments reveal that fusing textual guidance with the deepest and most abstract visual features (f\textsubscript{4}) significantly outperforms fusion at shallower levels. We find that more abstract visual features share a higher degree of semantic density with textual features, which we believe is a key factor contributing to the observed performance improvement. Motivated by this finding, we further investigate the role of semantic aggregation in our subsequent experiments. We observe that during domain generalization, the model’s inner feature representations tend to become dispersed---pixels belonging to the same semantic category (e.g., left ventricle) are widely scattered in the feature space, lacking compactness\cite{gao2022pyramidclip}. Based on this observation, we propose that enhancing semantic aggregation can facilitate better alignment and fusion between the visual and textual modules. When semantic features are more tightly clustered around their semantic centers, the model can more accurately capture the core semantics shared between text and image while avoiding redundant peripheral information. This, in turn, helps reduce the loss of essential information during the multimodal alignment and fusion process, resulting in fused features within the shared space that are more closely aligned with their underlying semantics.

To improve semantic aggregation, we propose a novel framework that explicitly addresses semantic aggregation in both the alignment and fusion stages. The core idea is to first perform semantic aggregation on both textual and visual features, then align the processed textual features with high-level visual representations, and finally decode the fused features using joint information from both modalities. This process ensures semantic compactness and effectively mitigates the problem of dispersed feature representations. By achieving higher semantic compactness through the aggregation process, feature representations become more accurate, thereby enhancing the model’s semantic generalization capability. Our model achieves outstanding performance across multiple datasets.The main contributions of this work can be summarized as follows:

\begin{itemize}

    \item For the feature alignment stage, we design a novel Expectation-Maximization Attention mechanism combined with a deep fusion strategy. The former dynamically clusters dispersed pixel features into compact semantic centers, while the latter aligns textual features with high-level visual representations. The combination of these two mechanisms effectively enhances semantic compactness, resulting in more concentrated features that exhibit stronger semantic consistency, thereby significantly improving cross-domain alignment accuracy.

    \item For the feature fusion stage, we design a novel Text-Guided Pixel Decoder. Building upon visual transformer foundation model pre-trained with DINOv2\cite{Oquab_Darcet_Moutakanni_Vo_Szafraniec_Khalidov_Fernandez_Haziza_Massa_El-Nouby_etal.}, we introduced a parallel text-guidance stream, forming a text-guided dual-stream fusion learning framework. The Text-Guided Pixel Decoder plays a central role within this framework, enabling the fusion of features from the textual and visual domains. The clustered dual-stream features result in more precise and concentrated fused representations when projected into the shared space, achieving inter-modality complementarity while mitigating interference from redundant information.
    
    \item We conducted comprehensive experiments on public datasets of two distinct anatomical sites (cardiac and fundus). The results demonstrate that our method achieves significant performance improvements in testing, surpassing SOTA methods, including Rein\cite{Wei_2024_CVPR}.
\end{itemize}

\section{Related works}

\subsection{Domain Generalization for Medical Imaging}
\label{subsec:sdg_medical}

Domain Generalization (DG), which aims to learn a model from a source domain that generalizes well to unseen target domains, is a critical challenge in medical image analysis. Existing techniques primarily fall into two categories. Image-level methods, such as SLAug, attempt to diversify the source data through augmentations like style transfer, but they often only address low-level variations and fail to handle deeper, content-related domain shifts. On the other hand, feature-level methods aim to learn domain-invariant representations, often through feature disentanglement as seen in CSDG\cite{Ouyang_Chen_Li_Li_Qin_Bai_Rueckert} and CCSDG\cite{Hu_Liao_Xia_2023}, which separate style from content. However, a key limitation of these vision-only disentanglement approaches is that the learned "content" features, while style-agnostic, often lack explicit high-level semantic meaning. This deficiency can limit their robustness when faced with significant anatomical variations in new domains. Our work addresses this fundamental gap by integrating explicit textual semantic guidance, aiming to learn content features that are not only domain-invariant but also semantically rich.

\subsection{Cross-modal Learning for Segmentation}
\label{subsec:cross_modal}

Cross-modal learning, through Vision-Language Models (VLMs) like CLIP\cite{Radford_Kim_Hallacy_Ramesh_Goh_Agarwal_Sastry_Amanda_Mishkin_Clark_etal._2021}, has inspired a new wave of Text-Guided Image Segmentation methods\cite{jiang2025mmsdg}. These approaches generally follow two main strategies. The first, exemplified by DenseCLIP\cite{rao2021denseclip}, focuses on aligning pixel-level visual features with text embeddings, often by computing a per-pixel similarity map. The second, seen in models like TQDM\cite{pak2024textual}, leverages text embeddings as dynamic queries within a Transformer decoder to actively group corresponding pixel regions. Both strategies have demonstrated that language can serve as a powerful high-level prior to guide the learning of visual features.

However, these methods exhibit significant limitations when applied to specialized domains like medical imaging. The pixel-alignment approach often struggles because the fine-grained visual features of medical images do not perfectly align with the VLM's general-domain feature space. Similarly, the query-based approach's effectiveness degrades when faced with the subtle inter-class variations of medical images, as the predefined textual queries can become ambiguous. In essence, both strategies suffer from a persistent semantic gap between general-purpose language representations and domain-specific visual cues, and they lack a mechanism to prepare the features for a robust cross-modal interaction. Our work addresses this core challenge by introducing a novel mechanism to actively aggregate, fuse, and align both visual and textual features, bridging this gap to ensure a more effective fusion process.
\subsection{Attention Mechanism}
\label{subsec:attention_opt}

The self-attention mechanism is a key contributor to the success of Transformers\cite{vaswani2017attention}, but it also has limitations. In particular, when processing high-resolution medical images, its quadratic computational complexity can lead to inefficiency. Moreover, standard self-attention can be susceptible to irrelevant information when feature representations are noisy or dispersed.

To address these issues, various strategies have been proposed to optimize attention mechanisms. Some approaches combine clustering with attention by first grouping tokens and then computing attention within or between clusters, aiming to reduce complexity and enhance robustness. The Expectation-Maximization (EM) algorithm\cite{477e7e2b-4ded-3369-981e-9b40850a2701}, a classical unsupervised clustering method, provides a theoretical foundation for this goal.

Recently, works such as Mamba\cite{zhang2025mamba} have innovatively applied the EM algorithm to the design of attention mechanisms. Through iterative optimization, EM enables dynamic grouping and refinement of features, demonstrating great potential in tackling complex feature modeling challenges.

While our work is inspired by this direction of integrating EM principles into attention mechanisms, our approach differs significantly from Mamba\cite{zhang2025mamba} in both application scope and architectural design. Mamba\cite{zhang2025mamba} primarily utilizes EM-like state space models to create a new, general-purpose backbone architecture intended as a replacement for standard Transformers\cite{vaswani2017attention}. 

In contrast, our EM-Aggregation mechanism is not a backbone replacement, but rather a lightweight, plug-and-play refinement module specifically designed to address the challenges of multimodal domain generalization. Instead of redesigning the entire network, we strategically inject our EM-Aggregation modules as adapters into a powerful pre-trained visual backbone (ReinsDINOv2) and a parallel text encoder. This allows us to leverage the immense knowledge of existing foundation models while precisely targeting the feature dispersion problem at both the visual and textual levels. Therefore, our work focuses on using EM as a targeted tool for feature recalibration and robust fusion within a multimodal framework, rather than for general-purpose sequence modeling.

\section{Method}

\subsection{Problem Definition and Framework Overview}
\label{subsec:problem_and_framework}

\subsubsection{Problem Definition}
In the context of domain generalization, we define the source training set and the target set as follows:
\begin{equation}
D_S = \{(x_{i,S}, y_{i,S})\}_{i=1}^{N_S} \quad \text{and} \quad D_T = \{(x_{j,T}, y_{j,T})\}_{j=1}^{N_T}
\end{equation}
where $(x_{i,S}, y_{i,S})$ represents the $i$-th sample and its corresponding label from the source domain $D_S$, and $(x_{j,T}, y_{j,T})$ represents a sample and label from the target domain $D_T$. During the domain generalization process from $D_S$ to $D_T$, features learned by the model often become more dispersed and less accurate. Through extensive observation and study, we have identified that this performance degradation primarily stems from two issues: the model's insufficient understanding of the inherent semantics of the target, and inconsistencies in feature distribution caused by style variations.
\subsection{Motivation and Overview}
While the integration of semantic guidance from Vision-Language Models (VLMs) has shown great promise, existing methods like TQDM\cite{pak2024textual} still exhibit significant limitations when applied to the medical imaging domain. We argue that this performance gap stems from two fundamental challenges. Firstly, a persistent semantic gap exists between the general-domain language representations from VLMs and the highly specific, fine-grained visual features of medical images. Secondly, how to effectively leverage these semantic features to guide the visual backbone—without introducing noise or causing feature misalignment—remains a major hurdle.

To address these challenges, we propose a Text-Guided Dual-Stream Fusion Learning Framework. Based on a foundation model, this framework enhances semantic compactness from two key perspectives-mutimodal feature alignment and feature fusion-to effectively mitigate the issue of feature dispersion.

In the feature alignment component, we perform Expectation-Maximization (EM) methods on both visual and textual features. This process dynamically clusters the dispersed features around their most semantically relevant centers, preserving core semantics while suppressing redundant information. Moreover, we observe that textual features exhibit higher semantic similarity and comparable semantic compactness with high-level visual features, making them well-suited for alignment and fusion.

In the feature fusion component, the semantic guidance module integrates features from images and their corresponding textual descriptions within a unified semantic space, enabling the text embeddings to capture domain-invariant semantic information. This significantly improves segmentation accuracy and enhances the model's robustness in cross-domain tasks.

\begin{figure*}[htbp]
    \centering
    \includegraphics[width=\textwidth]{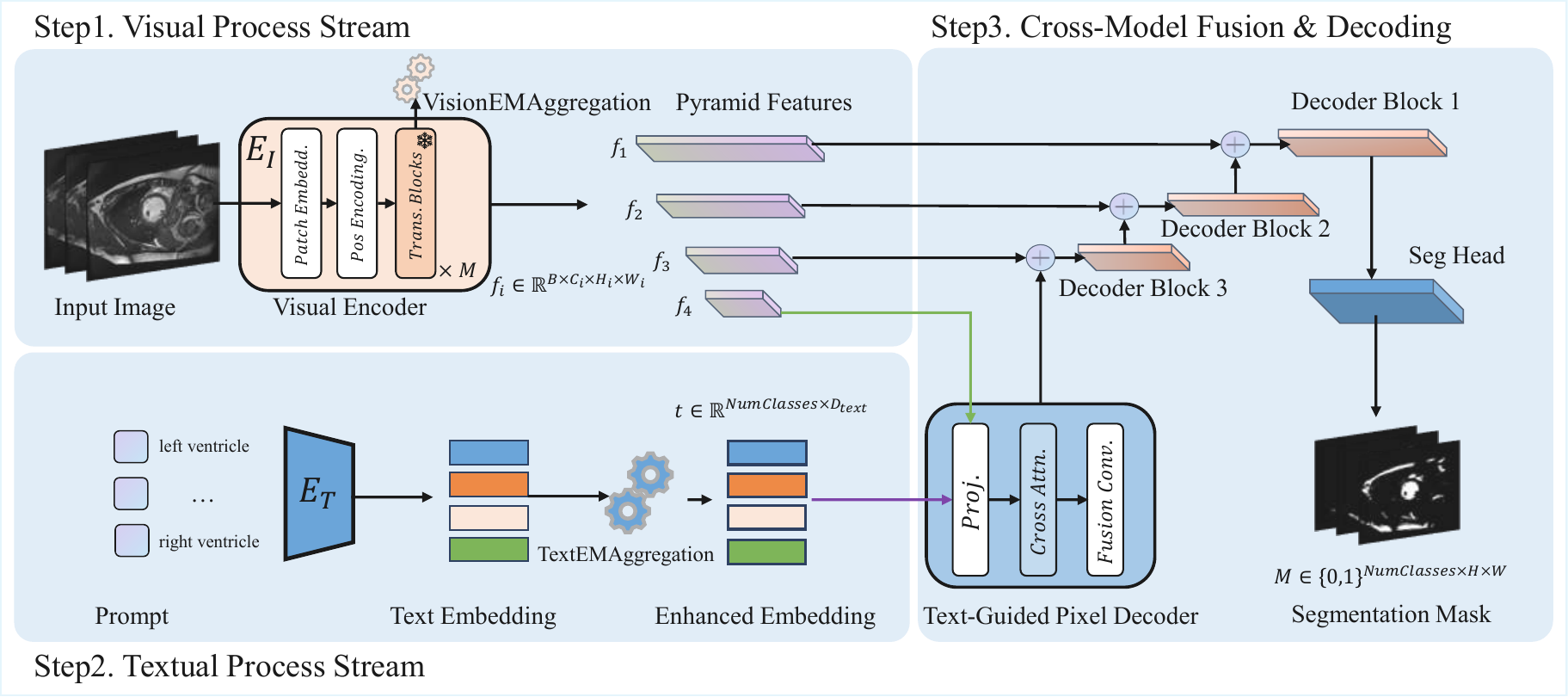} 
    \caption{The overall architecture of our proposed text-guided framework with EM. The framework consists of parallel visual and textual streams, which are deeply integrated within the decoder through our designed cross-modal fusion mechanism. }
    \label{fig:framework}
\end{figure*}

\begin{figure}[htbp]
    \centering
    \includegraphics[width=\columnwidth]{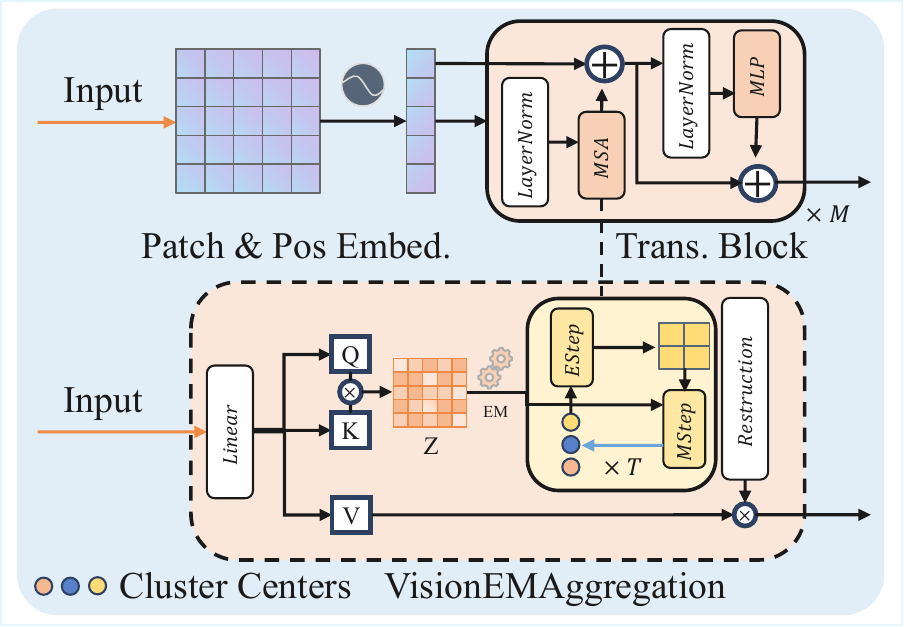} 
    \caption{The overall architecture of visual encoder. The framework consists of a progress of patch embedding and position embedding, and several transformer blocks }
    \label{fig:EV}
\end{figure}

\begin{figure}[htbp]
    \centering
    \includegraphics[width=\columnwidth]{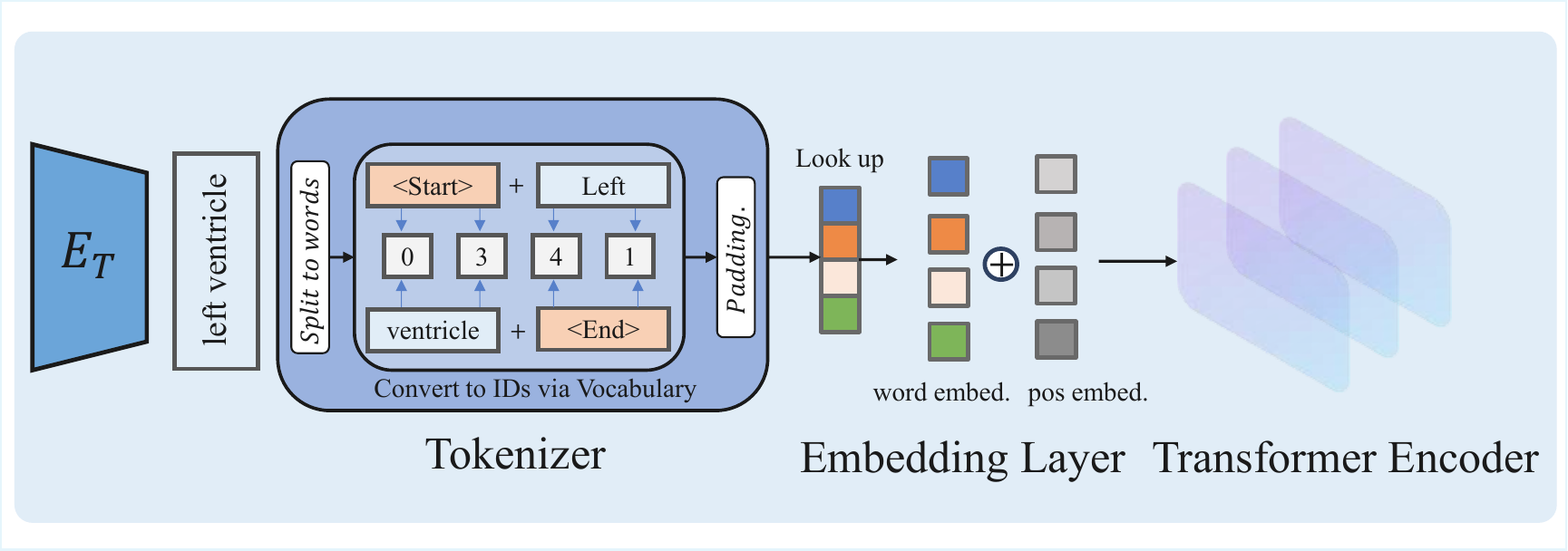} 
    \caption{The overall architecture of textual encoder. The framework consists of a tokenizer, an embbeding layer and several transformer encoders }
    \label{fig:ET}
\end{figure}

The overall architecture of our proposed text-guided framework with EM is illustrated in Fig.~\ref{fig:framework}. The framework follows a dual-stream design, primarily composed of three core components: a Visual Process Stream, a parallel Textual Process Stream, and a Cross-Modal Fusion \& Decoding module.\\
Our detailed visual encoder and textual encoder are illustrated in Fig.~\ref{fig:EV} and Fig.~\ref{fig:ET}. The visual encoder (Fig.~\ref{fig:EV}) processes the input image by first converting it into a sequence of patch and positional embeddings, which are then fed into a series of Transformer blocks. Crucially, we introduce our \texttt{VisionEMAggregation} mechanism to refine the visual features iteratively, enhancing their semantic coherence. The textual encoder (Fig.~\ref{fig:ET}) transforms raw text prompts into robust semantic embeddings through a pipeline of tokenization, an embedding layer, and a Transformer encoder, preparing them for the subsequent cross-modal fusion.

\subsection{Multimodal Alignment Strategy}
\label{subsec:alignment_strategy}

A central challenge in our framework is the effective alignment of features from disparate modalities: the low-dimensional, high-level semantics of text and the high-dimensional, detailed representations of vision. To address this, we propose a two-stage alignment strategy: first, an intra-modality feature refinement stage using our novel Expectation-Maximization Aggregation, followed by an inter-modality deep alignment stage.

\subsubsection{Expectation-Maximization Aggregation}
\label{subsubsec:em_aggregation}

To mitigate feature dispersion and facilitate a more effective cross-modal fusion, we introduce Expectation-Maximization Aggregation (EM-Aggregation). This mechanism refines features within each modality by modeling them as a mixture of $K$ latent semantic concepts, represented by prototypes $\bm{\mu}$. Given a set of input features $\mathbf{X} = \{\mathbf{x}_n\}_{n=1}^N$, the mechanism iteratively performs two steps:

E-step (Expectation): We compute the responsibility $z_{nk}$, the soft assignment of each feature $\mathbf{x}_n$ to each prototype $\bm{\mu}_k$, using the softmax function:
\begin{equation}
        z_{nk} = \frac{\exp(\mathbf{x}_n \cdot \bm{\mu}_k / \tau)}{\sum_{j=1}^{K} \exp(\mathbf{x}_n \cdot \bm{\mu}_j / \tau)}
\end{equation}
    
M-step (Maximization): We update the prototypes by computing the weighted average of the features, and then reconstruct the features based on the new prototypes:
\begin{equation}
        \bm{\mu}_k^{\text{new}} = \frac{\sum_{n=1}^{N} z_{nk} \mathbf{x}_n}{\sum_{n=1}^{N} z_{nk}}
\end{equation}
\begin{equation}
        \mathbf{x}_n^{\text{refined}} = \sum_{k=1}^{K} z_{nk} \bm{\mu}_k^{\text{new}}
\end{equation}
        
This iterative process serves a dual purpose: it not only alleviates dispersion by aggregating features into a more compact distribution but also produces more discriminative representations that are better suited for subsequent alignment\cite{jamshidian1997acceleration}. We apply this universal mechanism as two specialized instances. For the textual stream, Textual EM-Aggregation (T-EMA) treats the text embeddings as initial anchors and aggregates them into core meta-semantic centers, producing a robust set of semantic targets. Complementarily, for the visual stream, Visual EM-Aggregation (V-EMA), implemented as \texttt{VisionEMAggregation}, clusters deep patch features around latent concepts, removing domain-specific noise while preserving core semantic structures for a more robust fusion.

\subsubsection{Deep Alignment Strategy}
\label{subsubsec:deep_alignment}

Once the features from both the visual and textual streams have been independently refined by our EM-A mechanism, the next crucial step is to perform the actual cross-modal fusion. This is achieved through our Deep Alignment Strategy, which is implemented within the Text-Guided Pixel Decoder. This module serves as the primary locus for inter-modality interaction, taking the refined deep visual features and the refined semantic prototypes as input.

A critical design choice within this alignment stage is determining the optimal level at which to fuse the textual guidance with the visual feature pyramid \{f\textsubscript{1}, f\textsubscript{2}, f\textsubscript{3}, f\textsubscript{4}\}\cite{gao2022pyramidclip}. As empirically validated by our preliminary experiments (see Fig.~\ref{fig:fusion_level_ablation} in Introduction), fusion at deeper levels is significantly more effective. This is because deep features (e.g., f\textsubscript{4}) are more abstract and semantically rich, making them better suited for alignment with high-level textual concepts. Shallow features (e.g., f\textsubscript{1}), while rich in spatial detail, lack the semantic abstraction necessary for a meaningful cross-modal fusion. Therefore, based on this evidence, our framework strategically performs the text-guided fusion exclusively at the deepest feature level, f\textsubscript{4}, to maximize the effectiveness of the semantic guidance.

\subsection{Multimodal Fusion Strategy}
\label{subsec:med_seg}
The primary objective of the module is to provide text guidance and supervision for the image segmentation task\cite{fischer2024prompt,huang2024cat}. The main steps include:

First, encode the required domain-invariant semantic knowledge to the greatest extent possible to generate textual object queries.

Second, fuse the semantic guidance module with high-level features repeatedly extracted from the image to perform pixel-level feature enhancement.

Finally, apply regularization to the text-guided module to ensure that the textual component remains aligned with its original semantics.

\subsubsection{Text Query Generation}
To enable our deep alignment strategy, we first need to generate high-quality, domain-invariant semantic vectors, which we term Text Queries. This process begins with our vocabulary module, specifically designed for medical terminology to ensure precise semantic recognition. To harness the robust, pre-trained knowledge from the Vision-Language Model, we keep the core parameters of our text encoder, $E_T$, frozen. This preserves the domain-invariant nature of the learned semantics.

However, to adapt these general-purpose embeddings for the specific downstream task of segmentation, we introduce a set of learnable vectors known as prompts, denoted as $\mathbf{p}$. These prompts are prepended to the tokenized class names and are the only part of the text pipeline that is fine-tuned. This task-specific adaptation allows the model to learn contextual nuances for segmentation without disrupting the frozen encoder's core knowledge. The final text query $\mathbf{t}_c$ for a class $c$ is thus generated as:
\begin{equation}
    \mathbf{t}_c = E_T([\mathbf{p}, \text{tokenize}(c)])
    \label{eq:text_query_generation}
\end{equation}
Furthermore, to handle unknown words or variations in medical terms, our tokenizer incorporates a fault-tolerant mechanism based on partial matching, ensuring robustness against out-of-vocabulary inputs.We combine this module with the foudation model. In this way, these generated text queries, now both robust and task-adapted, are then passed to the Text-Guided Pixel Decoder for cross-modal fusion.

\subsubsection{Text-guided Pixel Decoder Mechanism}

\begin{figure}[htbp]
    \centering
    \includegraphics[width=\columnwidth]{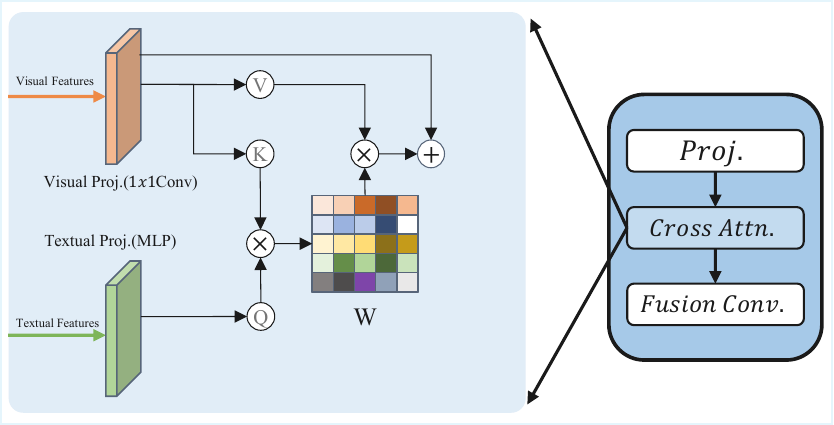} 
    \caption{The overall architecture of our text-guided pixel decoder. The framework consists of Projection layers (Proj), a Cross-Attention module (QKV), and a Fusion Convolutional block (Fusion Conv).}
    \label{fig:cross}
\end{figure}

To effectively leverage semantic information from text, we designed a mechanism to project both textual and medical image features into a shared embedding space. The overall architecture of this mechanism is illustrated in Fig.~\ref{fig:cross}. Through a cross-modal attention process, our model enhances the semantic aggregation of image features at the pixel level. This enables the textual guidance to refine visual representations, ensuring that pixels are expressed precisely based on their inherent semantics.

Specifically, we first project the textual features into a new space---the text-visual semantic space---using a multilayer perceptron (MLP) composed of linear layers and activation functions. Meanwhile, a \(1 \times 1\) convolution\cite{wei2020multi} is applied to adjust the channel dimension of the image features to match that of the text. After flattening the visual feature map, the model performs cross-modal attention matching between the image and text features.

The cross-modal attention mechanism we adopt is similar in nature to self-attention\cite{yuan2024hierarchical} in the foundation models, with the key difference being that the queries (Q) are derived from the textual features, while the keys (K) and values (V) come from the visual features. Specifically, under the multi-head attention setup, each head operates in an independent subspace with its own learned query, key, and value projections, followed by linear transformations.
\begin{equation}
\begin{aligned}
\mathbf{Q}_h = \mathbf{Q} \mathbf{W}_h^Q, \quad
\mathbf{K}_h = \mathbf{K} \mathbf{W}_h^K, \quad
\mathbf{V}_h = \mathbf{V} \mathbf{W}_h^V 
\end{aligned}
\end{equation}
Next, we perform attention computation for each head to measure the similarity between the queries and keys, and use the resulting scores to weight the value vectors as follows:
\begin{equation}
A^h = \text{Softmax}\left( \frac{Q^h {K^h}^\top}{\sqrt{d^h}} \right)
\end{equation}
\indent $\mathbf{A}_h \in \mathbb{R}^{B \times N_q \times N_k}$ represents the attention weight matrix. We apply a scaling factor of $\sqrt{d_h}$ to prevent excessively large dot products, which could lead to vanishing or exploding gradients. The value $A^h$ reflects the degree of attention each text query pays to every pixel location in the image and is used to compute and update pixel values. By doing so, we achieve semantic-centric clustering, where image regions belonging to the same semantic category are clustered together, enhancing feature representation capability.

Finally, we use the attention weights to compute weighted sums of the value vectors, obtaining the fused features for each query\cite{Khan_Naseer_Hayat_Zamir_Khan_Shah_2022}. Specifically, we compute $O^h = A^h V^h$, resulting in $\mathbf{O}^h \in \mathbb{R}^{B \times N_q \times d_h}$. The outputs from all attention heads $O^h$ are then concatenated and passed through an output layer to integrate the information:

\begin{equation}
O = \text{Concat}\left(O^1, \ldots, O^H \right) W^O
\end{equation}

This output fuses complementary information from multiple subspaces, capturing cross-modal correspondences from various perspectives, scales, and spatial locations.\\
\indent Finally, the attention results are projected back into the image feature space. All the visual information guided by the different text categories is aggregated into a unified, global text-guided context vector. This vector is then transformed into the visual feature modality and fused with the original projected visual features, resulting in a feature map that combines high-level visual representations with strong semantic discriminability.

\subsection{Implementation Details and Optimization}
\label{subsec:implementation}

\subsubsection{Loss Function Design}
Our model is trained using a composite loss function that combines pixel-wise and region-based supervision. The total segmentation loss, $\mathcal{L}_{\text{seg}}$, is a weighted sum of the Binary Cross-Entropy (BCE) loss and the Dice\cite{Milletari_Navab_Ahmadi_2016} loss:
\begin{equation}
    \mathcal{L}_{\text{seg}} = \lambda_{\text{bce}}\mathcal{L}_{\text{bce}} + \lambda_{\text{dice}}\mathcal{L}_{\text{dice}}
    \label{eq:compound_loss}
\end{equation}
where $\mathcal{L}_{\text{bce}}$ provides fine-grained, per-pixel supervision, which is beneficial for learning precise boundaries. Concurrently, $\mathcal{L}_{\text{dice}}$ directly optimizes the Dice score, offering robustness against class imbalance. The total BCE and Dice losses are aggregated across all target classes (e.g., LV, Myo, RV).

\subsubsection{Textual Regularization}
To ensure that the learned text prompts\cite{zhou2022learning} remain semantically stable and discriminative throughout training\cite{cui2025similarity}, we introduce a textual regularization loss, $\mathcal{L}_{\text{reg}}^{\text{lang}}$. Let $\mathbf{T} \in \mathbb{R}^{C \times d}$ be the text embeddings generated by the trainable text encoder and $\mathbf{R} \in \mathbb{R}^{C \times d}$ be a set of pre-defined, fixed anchor embeddings. The regularization loss is defined as the cross-entropy between the softmax of their similarity matrix and an identity matrix $\mathbf{I}_C$:
\begin{equation}
    \mathcal{L}_{\text{reg}}^{\text{lang}} = \text{Cross-Entropy}\left( \text{Softmax}(\mathbf{T} \mathbf{R}^{\top} / \tau),\ \mathbf{I}_C \right)
    \label{eq:lang_reg}
\end{equation}
where $\tau$ is a temperature parameter. This contrastive formulation encourages the embedding of each class to be most similar to its corresponding anchor feature, effectively preventing semantic drift by pulling positive pairs closer while pushing negative pairs apart in the feature space.

\subsubsection{Total Loss and Optimization}
The final objective function, $\mathcal{L}_{\text{total}}$, combines the segmentation and regularization losses:
\begin{equation}
    \mathcal{L}_{\text{total}} = \mathcal{L}_{\text{seg}} + \lambda_{\text{reg}}\mathcal{L}_{\text{reg}}^{\text{lang}}
\end{equation}
where $\lambda_{\text{bce}}$, $\lambda_{\text{dice}}$, and $\lambda_{\text{reg}}$ are empirically set weighting factors(In this research, we set them as 1.0, 0.3, 0.05). We employed the Adam\cite{kingma2014adam} optimizer for end-to-end training. The initial learning rate was set to $1 \times 10^{-4}$ with a weight decay of $10^{-4}$. A learning rate scheduler was used to smoothly anneal the learning rate. To enhance training efficiency and stability, we utilized Automatic Mixed Precision (AMP) and gradient clipping with a max L2-norm of 1.0.

\section{Experiments}
\label{sec:experiments}

\subsection{Datasets and Evaluation Metrics}
\label{subsec:datasets_metrics}

Our experiments were conducted on two public, multi-center medical image segmentation datasets, representing different anatomical sites and imaging modalities.

For the cardiac task, we employed T2-weighted MRI images from the publicly available ACDC dataset~\cite{8360453}.The combined dataset comprises a total of 1479 T2-weighted MRI volumes from five distinct domains, including 626 ACDC-bSSFP, 121 MSCMR-bSSFP, 49 MSCMR-T2, 448 MSCMR-LGE, and 235 EMIDEC-DEMRI. In our experimental setup, we designated the data from ACDC-bSSFP as the SOURCE domain for training. The data from the remaining four vendors were then used as strictly unseen TARGET domains to evaluate the model's generalization ability under complex domain shifts. All samples were resized to $224 \times 224$ and normalized to have zero mean and unit variance.

For the fundus task, we utilized the FairDomain dataset~\cite{tian2024fairdomain}, which comprises 8000 Scanning Laser Ophthalmoscopy (SLO) and 8000 Optical Coherence Tomography (OCT) images from multiple centers for the segmentation of the optic cup (OC) and optic disc (OD). This dataset is well-suited for domain generalization studies due to the significant modality gap between SLO and OCT images. In our experiments, we conducted two sets of evaluations: one where the SLO images were designated as the SOURCE domain and OCT as the TARGET domain, and vice versa. All images were resized to $224 \times 224$ for training and evaluation.

To quantitatively assess segmentation performance, we employed a widely used metrics:Dice-Sørensen Coefficient (DSC): Measures the overlap between the predicted segmentation and the ground truth label. It is calculated as $DSC = \frac{2|P \cap G|}{|P| + |G|}$, where P and G represent the predicted and ground truth regions, respectively.

We calculated these metrics for each target class and report their average values.

\subsection{Implementation Details}
\label{subsec:implementation_details}

Our model was implemented based on the PyTorch framework. The visual encoder employs DINOv2\cite{Oquab_Darcet_Moutakanni_Vo_Szafraniec_Khalidov_Fernandez_Haziza_Massa_El-Nouby_etal.} (ViT-Base) as its backbone. The text encoder consists of a 3-layer Transformer with a word embedding dimension of 256. For our EM-Aggregation modules, the number of clusters K for \texttt{TextEMAggregation} was set to 8 with 6 iteration stages. For \texttt{VisionEMAggregation}, K was set to 6 with 1 iteration stage, and the strength factor $\alpha$ was set to 0.1. All input images were resized to $224 \times 224$ resolution. We used the Adam\cite{kingma2014adam} optimizer for training with an initial learning rate of $1 \times 10^{-4}$ and a weight decay of $1 \times 10^{-4}$. The batch size was set to 8. All experiments were conducted on a single NVIDIA RTX 4090 GPU.

\subsection{Comparison with State-of-the-Art Methods}
\label{subsec:sota_comparison}

We compared our proposed framework (Ours) against a comprehensive set of competing methods on both the cardiac and fundus datasets. We included one trained on the target domain (Upper bound). Additionally, we considered several SOTA SDG methods, spanning both CNN-based and Transformer-based architectures. To ensure fair comparisons, all competing methods were re-implemented in our framework using the same backbone (U-Net\cite{10.1007/978-3-319-24574-4_28} or ViT\cite{Khan_Naseer_Hayat_Zamir_Khan_Shah_2022})

All our experiments adhere to a rigorous leave-one-domain-out protocol to ensure a fair and reliable evaluation. Specifically, hyperparameters were optimized via 3-fold cross-validation on the source domain. Subsequently, the model was trained on the complete source dataset with the optimal parameters for 100 epochs  to ensure convergence.\\ Finally, the generalization performance was evaluated on strictly unseen target domains. Throughout this process, we strictly followed the principle of data separation: target domain data was never used for training, hyperparameter tuning, or early stopping decisions.

The quantitative results are presented in Table~\ref{tab:sota_cardiac_full} (cardiac) and Table~\ref{tab:fundus_sota_dice_final} (fundus). The results indicate a significant improvement of our method over the baseline and consistently show superior performance against all competing SDG methods in terms of average Dice Score across both tasks. For the cardiac task, it is noteworthy that our method fundamentally improves the ViT\cite{Khan_Naseer_Hayat_Zamir_Khan_Shah_2022}-based baseline; the vision-only Rein\cite{Wei_2024_CVPR} model performs poorly, but with our text guidance and EM-Aggregation, the full model significantly surpasses other ViT\cite{Khan_Naseer_Hayat_Zamir_Khan_Shah_2022} competitors like H2Former\cite{10093768}. Similarly, for the fundus task, our model achieves the best performance in both the OCT$\rightarrow$SLO and the more challenging SLO$\rightarrow$OCT directions. These consistently leading results across different anatomies and imaging modalities strongly prove that our framework is a robust and widely applicable solution for domain generalization.

\begin{table}[htbp]
   {\fontfamily{lmr}\selectfont
  \centering
  \caption{Quantitative comparison with SOTA methods on the ACDC dataset\cite{8360453} (Dice Score). Models were trained on a single source domain (ACDC-bSSFP) and tested on four unseen target domains(The performance of upper bound method on Domain 2 is limited, which is expected due to its small sample size.). Our method is marked in \colorbox{lightblue}{blue}. The best and second-best results are highlighted in \textbf{bold} and \underline{underline}, respectively.}
  \label{tab:sota_cardiac_full}
  
  \resizebox{\textwidth}{!}{
  \begin{tabular}{l|ccc|ccc|ccc|ccc|ccc}
    \toprule
    \multirow{2}{*}{\textbf{Methods}} & \multicolumn{3}{c|}{\textbf{Domain 1}} & \multicolumn{3}{c|}{\textbf{Domain 2}} & \multicolumn{3}{c|}{\textbf{Domain 3}} & \multicolumn{3}{c|}{\textbf{Domain 4}} & \multicolumn{3}{c}{\textbf{Average}} \\
    \cmidrule(lr){2-4} \cmidrule(lr){5-7} \cmidrule(lr){8-10} \cmidrule(lr){11-13} \cmidrule(lr){14-16}
    & Dice\_LV & Dice\_Myo & Dice\_RV & Dice\_LV & Dice\_Myo & Dice\_RV & Dice\_LV & Dice\_Myo & Dice\_RV & Dice\_LV & Dice\_Myo & Dice\_RV & Dice\_LV & Dice\_Myo & Dice\_RV \\
    \midrule
    \multicolumn{16}{l}{\textit{Upper Bound (Trained on Target Domain)}} \\
    \midrule
    U-Net\cite{10.1007/978-3-319-24574-4_28} & 0.7818 & 0.5260 & 0.5084 & - & - & - & 0.8716 & 0.7374 & 0.9787 & 0.8943 & 0.7604 & 0.7753 & 0.8492 & 0.6746 & 0.7541 \\
    H2Former\cite{10093768} & 0.9265 & 0.7598 & 0.7942 & - & - & - & 0.8996 & 0.8059 & 0.9704 & 0.9483 & 0.8823 & 0.8655 & 0.9248 & 0.8160 & 0.8767 \\
    \midrule
    \multicolumn{16}{l}{\textit{CNN-based Methods}} \\
    \midrule
    U-Net(2015)\cite{10.1007/978-3-319-24574-4_28} & 0.8992 & 0.7407 & 0.7407 & 0.1428 & 0.1054 & 0.1164 & 0.2847 & 0.3045 & 0.5234 & 0.7873 & 0.5824 & 0.5773 & 0.5285 & 0.4333 & 0.4895 \\
    CCSDG(2023)\cite{Hu_Liao_Xia_2023} & 0.9293 & 0.8019 & \underline{0.8304} & 0.4248 & 0.4434 & 0.2565 & 0.5343 & 0.5499 & 0.1531 & 0.8704 & \textbf{0.7129} & \textbf{0.7465} & 0.6897 & 0.6270 & 0.4966 \\
    CSDG(2021)\cite{Ouyang_Chen_Li_Li_Qin_Bai_Rueckert} & 0.8878 & 0.7075 & 0.7866 & 0.6423 & 0.4300 & \underline{0.4475} & \underline{0.7024} & \underline{0.6444} & 0.1021 & 0.8324 & 0.6486 & \underline{0.6859} & 0.7662 & 0.6076 & 0.5055 \\
    Cutout(2017)\cite{Devries_Taylor} & 0.9004 & 0.7381 & 0.7589 & 0.2773 & 0.2817 & 0.1304 & 0.5628 & 0.4571 & 0.5021 & 0.8094 & 0.6219 & 0.5713 & 0.6375 & 0.5247 & 0.4907 \\
    RandConv(2020)\cite{Xu_Liu_Yang_Raffel_Niethammer_2020} & 0.9183 & 0.7394 & 0.7718 & 0.1638 & 0.2946 & 0.0844 & 0.1879 & 0.2754 & \textbf{0.5736} & 0.7919 & 0.6171 & 0.5625 & 0.5155 & 0.4816 & 0.4981 \\
    MixStyle(2021)\cite{Zhou_Yang_Qiao_Xiang_2021} & 0.9075 & 0.7676 & 0.7778 & 0.1723 & 0.0529 & 0.0816 & 0.3879 & 0.3177 & 0.5017 & 0.8243 & 0.6088 & 0.6414 & 0.5730 & 0.4368 & 0.5006 \\
    SLAug(2022)\cite{Su_Yao_Yang_Wang_Sun_Huang_2022} & \textbf{0.9310} & \underline{0.8092} & 0.8123 & \underline{0.6480} & 0.4757 & \textbf{0.5089} & 0.6349 & 0.6230 & 0.0553 & \underline{0.8735} & 0.6207 & 0.6858 & \underline{0.7718} & \underline{0.6322} & \underline{0.5156} \\
    \midrule
    \multicolumn{16}{l}{\textit{ViT-based Methods}} \\
    \midrule
    Rein(2024)\cite{Wei_2024_CVPR} & 0.8617 & 0.3862 & 0.5959 & 0.5666 & 0.3041 & 0.1134 & 0.4616 & 0.3211 & \underline{0.5404} & 0.7998 & 0.2849 & 0.4069 & 0.6724 & 0.3241 & 0.4142 \\
    H2Former(2023)\cite{10093768} & \underline{0.9305} & \textbf{0.8229} & 0.7784 & 0.3546 & 0.3590 & 0.1370 & 0.3416 & 0.4151 & 0.2595 & 0.8263 & 0.6008 & 0.6158 & 0.6133 & 0.5495 & 0.4477 \\
    DAPSAM(2024)\cite{10.1007/978-3-031-72111-3_50} & 0.9004 & 0.7601 & 0.8205 & 0.4921 & 0.4437 & 0.1384 & 0.6443 & 0.5599 & 0.2297 & 0.8537 & 0.6673 & 0.6853 & 0.7226 & 0.6078 & 0.4685 \\
    tqdm(2024)\cite{pak2024textual} & 0.8899 & 0.5679 & 0.6836 & \textbf{0.6554} & \textbf{0.5679} & 0.3362 & 0.6769 & 0.4456 & 0.0596 & 0.8385 & 0.5422 & 0.6381 & 0.7652 & 0.5145 & 0.4294 \\
    \rowcolor{lightblue}
    \textbf{Ours} & 0.9284 & 0.8056 & \textbf{0.8313} & 0.5273 & \underline{0.5423} & 0.3891 & \textbf{0.7898} & \textbf{0.6972} & 0.2397 & \textbf{0.8745} & \underline{0.6854} & 0.6801 & \textbf{0.7800} & \textbf{0.6826} & \textbf{0.5351} \\
    \bottomrule
  \end{tabular}
   } 
   } 
\end{table}
\begin{table*}[htbp]
  {\fontfamily{lmr}\selectfont 
  \centering
  \caption{Ablation study on the Cardiac dataset. The source domain is ACDC-bSSFP, and the performance is evaluated on four unseen target domains. The table shows the Dice Score changes after progressively adding components. The best and second-best results are highlighted in \textbf{bold} and \underline{underline}, respectively.}
  \label{tab:ablation_acdc_full}
  
  \resizebox{\textwidth}{!}{
  
  \begin{tabular}{l|ccc|ccc|ccc|ccc|ccc}
    \toprule
    \multirow{2}{*}{\textbf{Method Configuration}} & \multicolumn{3}{c|}{\textbf{Domain 1}} & \multicolumn{3}{c|}{\textbf{Domain 2}} & \multicolumn{3}{c|}{\textbf{Domain 3}} & \multicolumn{3}{c|}{\textbf{Domain 4}} & \multicolumn{3}{c}{\textbf{Average}} \\
    \cmidrule(lr){2-4} \cmidrule(lr){5-7} \cmidrule(lr){8-10} \cmidrule(lr){11-13} \cmidrule(lr){14-16}
    & Dice\_LV & Dice\_Myo & Dice\_RV & Dice\_LV & Dice\_Myo & Dice\_RV & Dice\_LV & Dice\_Myo & Dice\_RV & Dice\_LV & Dice\_Myo & Dice\_RV & Dice\_LV & Dice\_Myo & Dice\_RV \\
    \midrule
    (1) Rein (Baseline) & 0.8617 & 0.3862 & 0.5959 & \textbf{0.5666} & 0.3041 & 0.1134 & 0.4616 & 0.3211 & 0.5404 & 0.7998 & 0.2849 & 0.4069 & 0.6724 & 0.3241 & 0.4142 \\
    (2) + Text & \underline{0.9281} & \textbf{0.8107} & 0.8249 & 0.4453 & 0.4816 & \underline{0.3081} & \underline{0.7706} & 0.6772 & \textbf{0.2681} & 0.8642 & \underline{0.6661} & 0.6381 & 0.7521 & \underline{0.6589} & 0.5098 \\
    (3) + Text EM & 0.9267 & 0.8052 & \underline{0.8296} & 0.4935 & \underline{0.4953} & 0.2991 & 0.7651 & \underline{0.6815} & 0.2341 & \underline{0.8661} & 0.6508 & \underline{0.6799} & \underline{0.7629} & 0.6582 & \underline{0.5107} \\
    (4) + Vision EM (Ours) & \textbf{0.9284} & \underline{0.8056} & \textbf{0.8313} & \underline{0.5273} & \textbf{0.5423} & \textbf{0.3891} & \textbf{0.7898} & \textbf{0.6972} & \underline{0.2397} & \textbf{0.8745} & \textbf{0.6854} & \textbf{0.6801} & \textbf{0.7800} & \textbf{0.6826} & \textbf{0.5351} \\
    \bottomrule
  \end{tabular}
  
  }
   } 
\end{table*}

\begin{table}[htbp]

  {\fontfamily{lmr}\selectfont 

  \centering
  \caption{Quantitative comparison of SOTA methods on the FairDomain\cite{tian2024fairdomain} fundus dataset (Dice Score). The improvement is statistically significant (p < 0.05) when compared to tqdm, using a Wilcoxon signed-rank test. Our method is marked in \colorbox{lightblue}{blue}. The best and second-best results are highlighted in \textbf{bold} and \underline{underline}, respectively.}
  \label{tab:fundus_sota_dice_final}
  \resizebox{\columnwidth}{!}{ 
  \begin{tabular}{ll cccc} 
    \toprule
    \multirow{2}{*}{\textbf{Method}} & \multirow{2}{*}{\textbf{Backbone}} & \multicolumn{2}{c}{\textbf{OCT $\rightarrow$ SLO}} & \multicolumn{2}{c}{\textbf{SLO $\rightarrow$ OCT}} \\
    \cmidrule(lr){3-4} \cmidrule(lr){5-6}
    & & OC (Cup) & OD (Disc) & OC (Cup) & OD (Disc) \\
    \midrule
    \multicolumn{6}{l}{\textit{Upper Bound}} \\
    \midrule
    U-Net\cite{10.1007/978-3-319-24574-4_28} & \multirow{1}{*}{CNN} & 0.8666 & 0.9511 & 0.8348 & 0.9274 \\
    H2Former\cite{10093768} & \multirow{1}{*}{ViT} & 0.8770 & 0.9595 & 0.8513 & 0.9343 \\
    \midrule
    \multicolumn{6}{l}{\textit{CNN-based Methods}} \\
    \midrule
    U-Net(2015)\cite{10.1007/978-3-319-24574-4_28} & \multirow{7}{*}{U-Net} & 0.7097 & 0.6656 & 0.4566 & 0.6837 \\
    Cutout(2017)\cite{Devries_Taylor} & & 0.4457 & 0.2618 & 0.5467 & 0.7811 \\
    RandConv(2020)\cite{Xu_Liu_Yang_Raffel_Niethammer_2020} & & 0.7429 & 0.7354 & 0.3772 & 0.6244 \\
    MixStyle(2021)\cite{Zhou_Yang_Qiao_Xiang_2021} & & 0.7200 & 0.7206 & 0.4395 & 0.7230 \\
    CSDG(2021)\cite{Ouyang_Chen_Li_Li_Qin_Bai_Rueckert} & & 0.7803 & 0.8185 & 0.6427 & 0.7993 \\
    SLAug(2022)\cite{Su_Yao_Yang_Wang_Sun_Huang_2022} & & 0.7325 & 0.7769 & 0.6501 & 0.8099 \\
    CCSDG(2023)\cite{Hu_Liao_Xia_2023} & & 0.7731 & 0.7744 & 0.6194 & 0.7867 \\
    \midrule
    \multicolumn{6}{l}{\textit{Transformer-based Methods}} \\
    \midrule
    Rein(2024)\cite{Wei_2024_CVPR} & & 0.7620 & \underline{0.8491} & 0.6454 & 0.8177 \\
    H2Former(2023)\cite{10093768} & \multirow{3}{*}{ViT} & 0.7725 & 0.7674 & 0.5561 & 0.8075 \\
    DAPSAM(2024)\cite{10.1007/978-3-031-72111-3_50} & & 0.7133 & 0.7450 & \underline{0.6947} & 0.8235 \\
    tqdm(2024)\cite{pak2024textual} & & \underline{0.7837} & 0.8305 & 0.6902 & \underline{0.8244} \\
    \rowcolor{lightblue}
    Ours\fnref{fn2}  & & \textbf{0.7862} & \textbf{0.8622} & \textbf{0.7034} & \textbf{0.8275} \\
    \bottomrule
  \end{tabular}
}
  }
\end{table}

\subsection{Ablation Studies}
\label{subsec:ablation_studies}

To investigate the individual contribution of each key component in our framework, we conducted a series of ablation studies. Starting with the vision-only Rein\cite{Wei_2024_CVPR} model as our baseline, we progressively integrated our core innovations: text guidance (+ Text Guidance), Textual EM-Aggregation (+ Text EM), and finally our complete framework which also includes Visual EM-Aggregation (Ours (Full Model)).

As demonstrated in Table~\ref{tab:ablation_acdc_full}, the results form a clear evidence chain of progressive performance improvement. The introduction of Text Guidance alone provides a substantial performance leap over the baseline, confirming the critical role of semantic priors in overcoming domain shift\cite{Reviewed_Blitzer_Crammer_Kulesza_Pereira_Vaughan_Wortman_Ben-David_Blitzer_Crammer_etal.}. Building upon this, the further addition of the Textual EM-Aggregation module yields another consistent performance boost, validating that refining the semantic features leads to more robust guidance. Finally, our Full Model, which integrates both textual and visual EM-Aggregation, achieves the best overall performance, demonstrating that our designed components are complementary and work in concert to achieve optimal domain generalization.

\subsection{Visualization and Qualitative Analysis}
\label{subsec:visualization}

To provide an intuitive demonstration of our model's capabilities, we present both qualitative segmentation comparisons and a mechanistic analysis of our EM-Aggregation module.

\begin{figure*}[htbp]
  \centering
  \includegraphics[width=\textwidth]{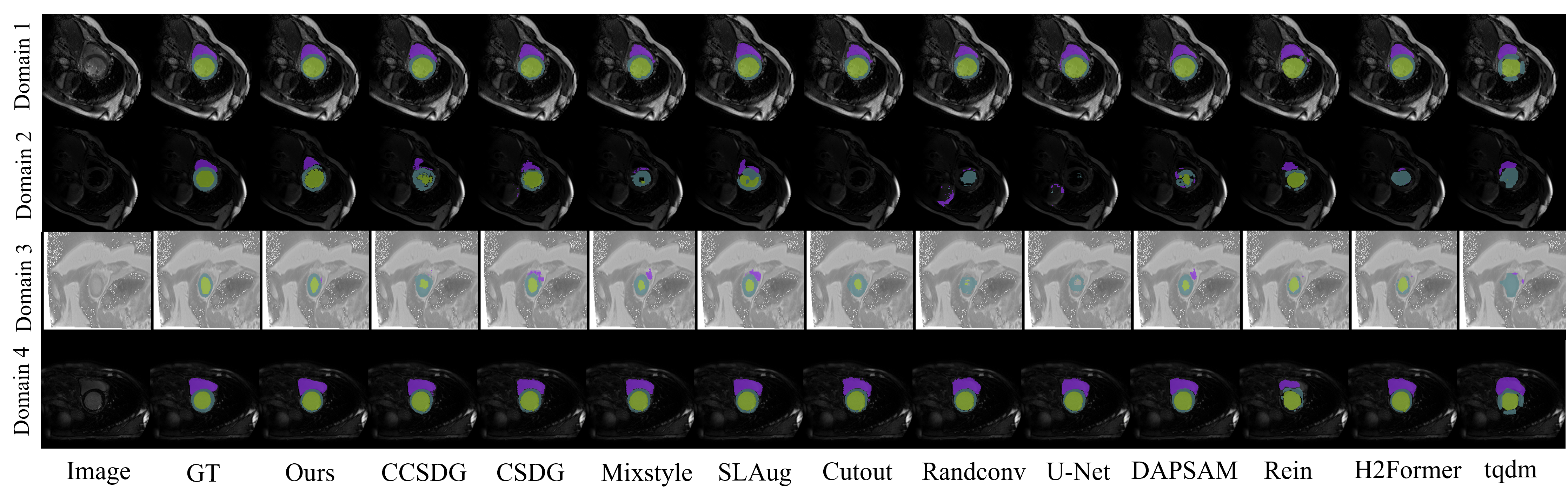}
  \caption{Visual comparison of segmentation results on challenging unseen samples from three different target domains of the ACDC dataset. Each row corresponds to a different target domain, and each column represents a different method. The left ventricle (LV) is shown in green, myocardium (Myo) in blue, and right ventricle (RV) in purple.}
  \label{fig:qualitative_cardiac}
\end{figure*}

\begin{figure*}[htbp]
  \centering
  \includegraphics[width=\textwidth]{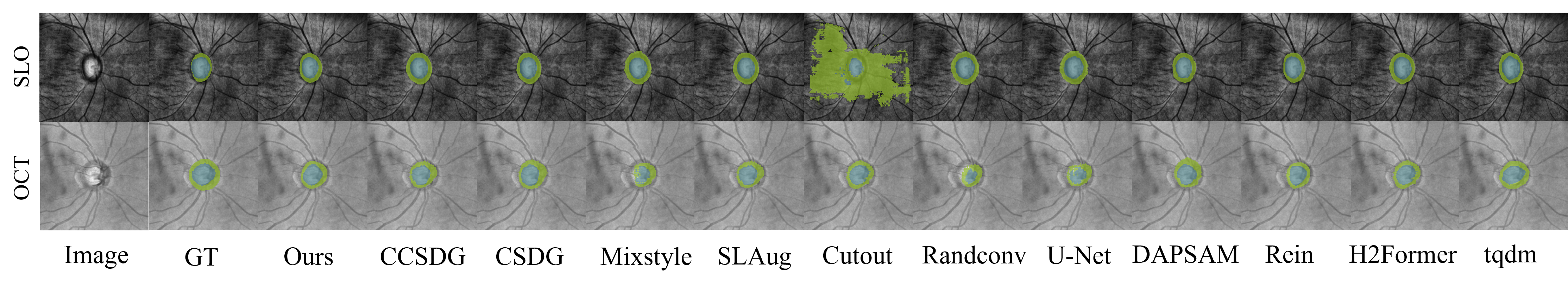}
  \caption{Visual comparison of segmentation results on representative unseen samples from the SLO and OCT domains. Each column displays the result of a different method. The optic disc is shown in green and the optic cup in blue.}
  \label{fig:qualitative_fundus}
\end{figure*}

Qualitative results on both the ACDC cardiac dataset (Fig.~\ref{fig:qualitative_cardiac}) and the FairDomain fundus dataset (Fig.~\ref{fig:qualitative_fundus}) visually corroborate our quantitative findings. Across all tested domains and modalities, our proposed method consistently produces segmentation masks that are more accurate, coherent, and anatomically plausible than competing state-of-the-art methods. The advantages are particularly pronounced in challenging cases with strong domain shifts or low image quality, where many baseline methods produce fragmented or erroneous predictions. In stark contrast, our framework demonstrates exceptional robustness, yielding clean and reliable segmentations.

\begin{figure}[htbp]
  \centering
  \includegraphics[width=\columnwidth]{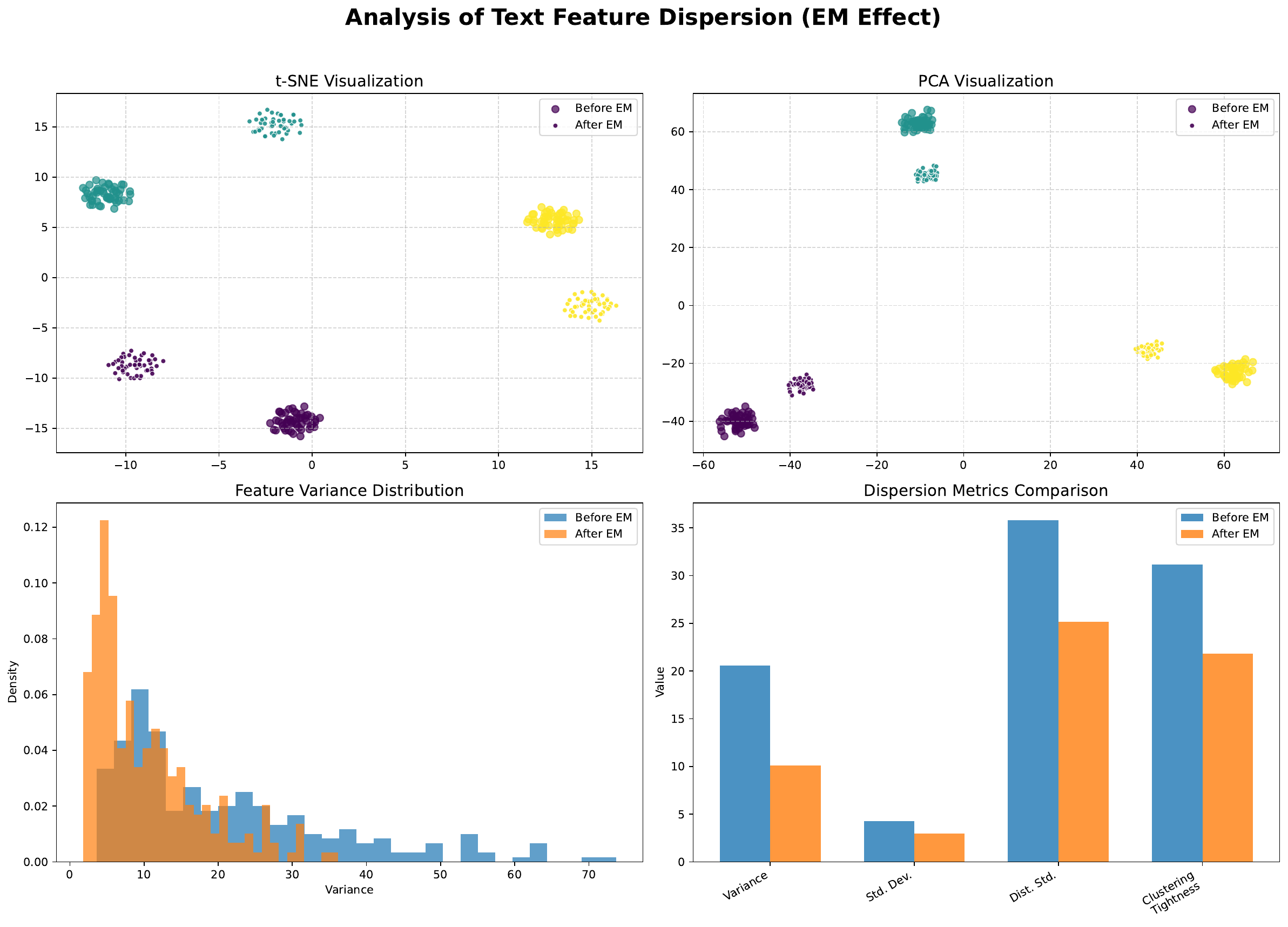} 
  \caption{Analysis of the effect of EM-Aggregation on text features. This figure compares the distributional changes of text embeddings before and after being processed by the EM mechanism.}
  \label{fig:text_vis}
\end{figure}

\begin{figure}[htbp]
  \centering
  \includegraphics[width=\columnwidth]{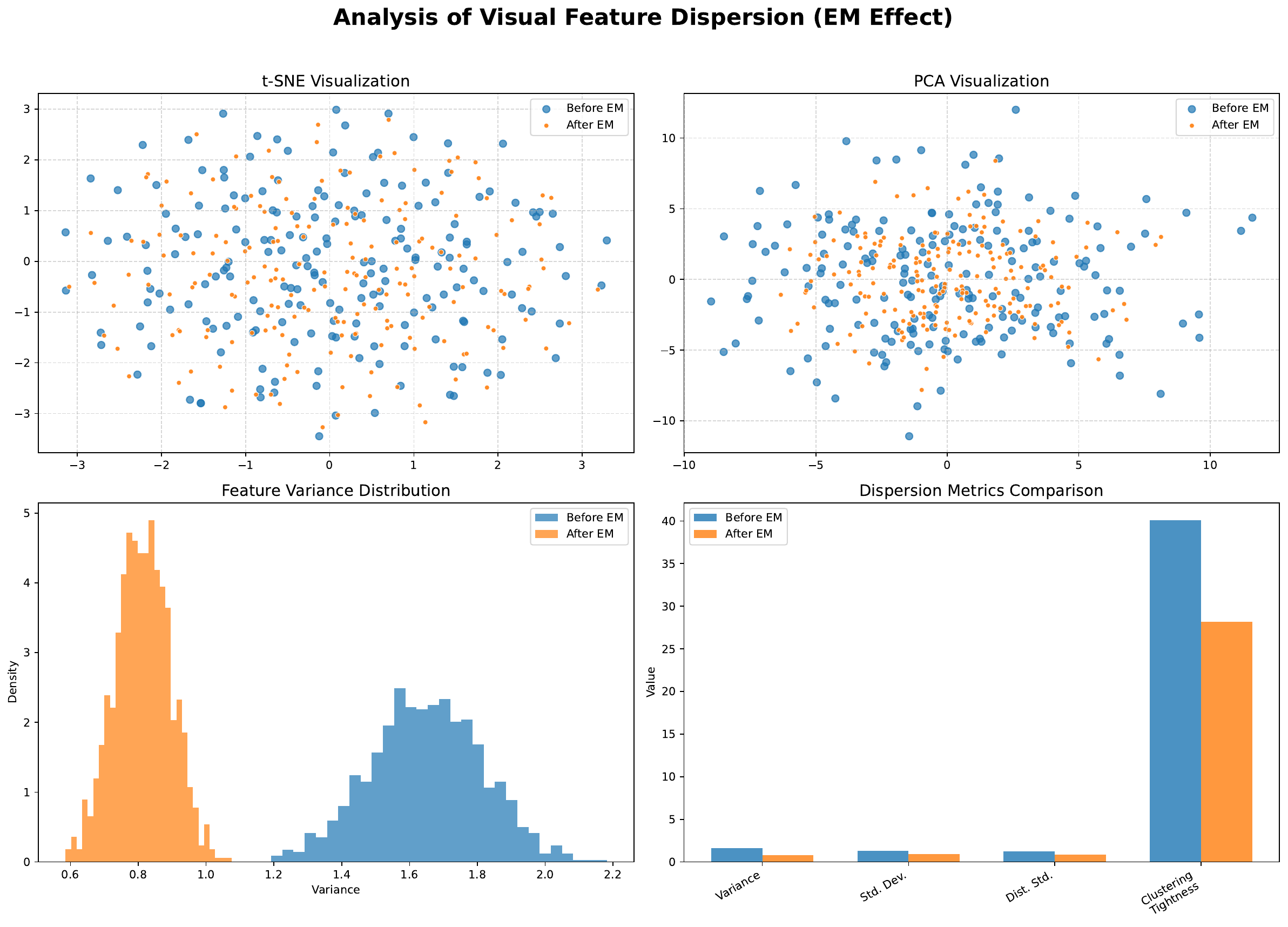}
  \caption{Analysis of the effect of EM-Aggregation on visual features. This figure illustrates the effect of the VisionEMAggregation module on patch token features. Echoing the results in Fig.~\ref{fig:text_vis}, the EM mechanism reduces the dispersion of visual features.}
  \label{fig:visual_vis}
\end{figure}

To mechanistically validate the effectiveness of our EM-Aggregation in mitigating feature dispersion, we analyzed the distributional changes of both textual and visual features. The results, presented in Fig.~\ref{fig:text_vis} and Fig.~\ref{fig:visual_vis}, reveal a consistent and significant compacting effect across both modalities. 

For the textual features (Fig.~\ref{fig:text_vis}), the \texttt{TextEMAggregation} module reduces the average feature Variance by a substantial 50.76\% (from 20.59 to 10.14) and improves Clustering Tightness by 30.00\% (from 31.18 to 21.82). Similarly, for the visual features (Fig.~\ref{fig:visual_vis}), the \texttt{VisionEMAggregation} module achieves a remarkable 50.71\% reduction in Variance (from 1.64 to 0.81) and a 29.79\% improvement in Clustering Tightness (from 40.12 to 28.17). These parallel findings strongly prove that our EM-Aggregation functions as a universal refinement engine, providing a solid, data-driven foundation for the model's superior generalization performance by learning more compact and robust representations.

\section{Discussion}
\label{sec:discussion}

Despite its strong performance, our framework presents avenues for future improvement. The current EM-Aggregation mechanism, while effective, introduces computational overhead that could be optimized using techniques like sparse attention, especially for high-resolution images. Additionally, its reliance on pre-defined class names may limit applicability to tasks with novel or difficult-to-describe lesions. Future work will focus on adapting our framework to a broader range of tasks, such as tumor segmentation, and investigating more sophisticated, adaptive fusion strategies to further enhance its robustness and clinical applicability.

\section{Conclusion}
\label{sec:conclusion}

In this paper, we addressed the challenging task of single-domain generalization in medical image segmentation. We introduced a novel framework founded on three synergistic principles: first, the integration of a text-guided semantic stream with a powerful visual transformer backbone to provide explicit, domain-invariant guidance. Second, the implementation of a deep fusion strategy, where we empirically demonstrated that aligning textual semantics with the most abstract visual features (f\textsubscript{4}) is critical for optimal performance. Finally, to facilitate this deep alignment, we proposed an Expectation-Maximization Aggregation (EM-Aggregation) mechanism. This module serves a dual purpose: it not only mitigates feature dispersion by refining representations into more compact distributions but, more importantly, promotes a more effective and robust fusion by producing dense and discriminative features for both modalities. Extensive experiments on both cardiac and fundus datasets confirm that this integrated approach significantly improves generalization performance, offering a promising direction for multimodal analysis in medical imaging.
\section{Acknowledgements}
		This work was supported by the Natural Science Foundation of Gansu Province, China under Grant 25JRRA666 and in part by the fundamental research funds for the central universities of China under
         Grant lzujbky-2023-it40, lzujbky-2024-it52 and in part by Key Laboratory of AI and Information Processing, Education Department of Guangxi Zhuang Autonomous Region (Hechi University), under Grant No. 2024GXZDSY004

\bibliographystyle{ieeetr}
\bibliography{refs}
\end{document}